%% file: main.tex
\documentclass[11pt]{article}

\usepackage[final]{acl}

\usepackage{times}
\usepackage{latexsym}
\usepackage[T1]{fontenc}
\usepackage[utf8]{inputenc}
\usepackage{microtype}
\usepackage{inconsolata}
\usepackage{graphicx}

\usepackage{booktabs}       
\usepackage{amsmath}        
\usepackage{amssymb}        
\usepackage{multirow}       
\usepackage{xcolor}         
\usepackage{subcaption}     
\usepackage{placeins}       
\usepackage{verbatim}       
\usepackage{listings}       
\usepackage{enumitem}      
\usepackage[most]{tcolorbox} 

\newtcolorbox{lawquote}[1][]{
  enhanced,
  colback=gray!4, colframe=gray!40, coltitle=black,
  boxrule=0.4pt, arc=2pt,
  left=10pt, right=10pt, top=8pt, bottom=8pt,
  fonttitle=\bfseries\small,
  attach boxed title to top left={xshift=10pt, yshift=-7pt},
  boxed title style={colback=white, colframe=gray!40, boxrule=0.4pt, arc=2pt, size=fbox, top=2pt, bottom=2pt, left=4pt, right=4pt},
  #1
}

\newtcolorbox{casebox}[1][]{
  enhanced,
  colback=orange!4, colframe=orange!35, coltitle=black,
  boxrule=0.4pt, arc=2pt,
  left=10pt, right=10pt, top=8pt, bottom=8pt,
  fonttitle=\bfseries\small,
  attach boxed title to top left={xshift=10pt, yshift=-7pt},
  boxed title style={colback=white, colframe=orange!35, boxrule=0.4pt, arc=2pt, size=fbox, top=2pt, bottom=2pt, left=4pt, right=4pt},
  #1
}

\newtcolorbox{promptbox}[1][]{
  enhanced,
  colback=gray!3, colframe=gray!40, coltitle=black,
  boxrule=0.4pt, arc=2pt,
  left=8pt, right=8pt, top=6pt, bottom=6pt,
  fonttitle=\bfseries\footnotesize,
  attach boxed title to top left={xshift=8pt, yshift=-6pt},
  boxed title style={colback=white, colframe=gray!40, boxrule=0.4pt, arc=2pt, size=fbox, top=2pt, bottom=2pt, left=4pt, right=4pt},
  #1
}

\lstdefinestyle{prompt}{%
  basicstyle=\footnotesize\ttfamily,
  breaklines=true,
  breakatwhitespace=true,
  breakindent=0pt,
  columns=fullflexible,
  keepspaces=true,
  showstringspaces=false,
  frame=none,
  xleftmargin=0pt,
  xrightmargin=0pt,
  postbreak=\mbox{\textcolor{gray}{$\hookrightarrow$}\space},
  literate=%
    {—}{{---}}{1}
    {–}{{--}}{1}
    {→}{{$\rightarrow$}}{1}
    {←}{{$\leftarrow$}}{1}
    {⇒}{{$\Rightarrow$}}{1}
    {’}{{'}}{1}
    {‘}{{`}}{1}
    {“}{{``}}{1}
    {”}{{''}}{1}
    {…}{{\ldots}}{1}
    {ü}{{\"u}}{1}
    {ö}{{\"o}}{1}
    {ä}{{\"a}}{1}
    {ß}{{\ss}}{1},
}

\definecolor{prov1}{HTML}{CC6677}   
\definecolor{prov2}{HTML}{332288}   
\definecolor{prov3}{HTML}{DDCC77}   
\definecolor{prov4}{HTML}{117733}   
\definecolor{prov5}{HTML}{88CCEE}   
\definecolor{prov6}{HTML}{882255}   
\definecolor{prov7}{HTML}{44AA99}   
\definecolor{prov8}{HTML}{999933}   
\definecolor{prov9}{HTML}{AA4499}   
\definecolor{prov10}{HTML}{888888}  
\newcommand{\provswatch}[1]{\textcolor{prov#1}{\rule{0.65em}{0.65em}}}

\title{By Their Fruits You Will Know Them: Comparing Formalizations of Law by the Decisions They Encode}

\author{Julius Vernie \\
  Technical University of Munich (TUM)\\
  \texttt{julius.vernie@tum.de} \\\And
  Matthias Grabmair \\
  Technical University of Munich \\
  \texttt{matthias.grabmair@tum.de} \\}

\begin{document}
\maketitle

\begin{abstract}
Formalizing legal provisions promises machine-accessible law and
automated legal reasoning, and recent LLMs make it tempting to
generate such formalizations directly from statutory text. However, any formalization makes implicit interpretive choices whose consequences are hard to 
anticipate, especially if an LLM is the author. We present a
method for systematically comparing different formalizations of the
same legal provision by their inferences on individual cases. Given multiple formalizations
of a provision, we match them at the node level, derive a shared
interface for each pair from the matching, and use a SAT solver to
enumerate the edge cases on which any two formalizations disagree.
Selected edge cases are then verbalized into concrete factual
scenarios that a legal expert can examine and act on. We apply our method 
to formalizations of ten EU provisions generated by nine frontier LLMs.
We find that structural alignment does not predict behavioral
divergence. Formalizations whose matching aligns most of their
structure still disagree on many cases.
The verbalized cases reveal qualitatively
distinct types of disagreement, including divergences that mirror
genuine controversies in the legal commentary.
\end{abstract}

\input{sections/introduction}
\input{sections/related_work}

\input{sections/data_set}

\input{sections/methods}
\input{sections/results}

\input{sections/conclusion}
\FloatBarrier  

\section*{Limitations}
Several caveats constrain our results. First, the pipeline depends
on matching nodes of different formalizations via an LLM, a step
whose reliability we can assess but whose correctness we cannot
(Section~\ref{sec:methods-matching-method}). We performed automated checks to 
assure the formal integrity of the matching results. 
In manual inspections we found the matchings to be mostly accurate, but we did sometimes observe nodes
placed in separate ECs that, in our opinion, should have been
matched.  This may be partly an artifact of our requirement that the
EC graph be loop-free.  Quantifying the effect would require expert
annotation of matchings, which we did not collect.

Second, we provide each LLM only the text of the provision to be
formalized. No surrounding regulatory context such as related articles,
recitals, official guidelines or case law is supplied. A legal
expert formalizing the same article would routinely consult these
sources to disambiguate terms, resolve cross-references, and check
the encoded reading against established interpretation. The quality
of our formalizations is therefore bounded by what each LLM already
encodes about EU law in its training, which is unevenly distributed
across instruments and known to under-represent recent or
specialized material.

Third, we formalize single articles rather than full
regulations. This keeps the task tractable but ignores the
inter-article structure that gives statutory law much of its
meaning: definitions in early articles (e.g.\ Art.~4 GDPR) propagate
throughout, exceptions in one article modify another, and
substantive provisions often depend on procedural conditions defined
elsewhere. Each article is treated as if it were self-contained.

Lastly, verbalization hides several layers of complexity. The
verbalization itself is generated by an LLM and may not faithfully
encode the underlying prime implicant; we do not separately evaluate
this fidelity, so an unfaithful scenario could elicit a verdict
unrelated to the actual disagreement. Furthermore, formalizations can themselves be legally incorrect but still conclude correct outcomes in some cases. A reviewer's agreement with the outcome on a single
scenario therefore is only weak evidence about the formalization as a whole.
Verbalization-based review is therefore one signal among many rather than
a definitive correctness test.

\section*{Ethical Considerations}

Our pipeline analyzes LLM-generated formalizations of real,
high-impact EU regulation.  These formalizations are not validated
for autonomous legal decision-making, and our central finding is that
they can disagree substantially with one another.  The pipeline is an
audit aid for expert review, not a component of an automated legal
system; a mis-formalization deployed without expert review could
misassign rights or liability.

The analysis identifies where formalizations diverge, not
which one is correct.  The per-model statistic of
Section~\ref{sec:results-edge-case-analysis} measures how often a
model's formalizations differ from those of the other models, and
must not be read as a quality ranking.

The verbalized scenarios concern fictional entities.  No personal
data was processed and no human subjects were involved.

Total LLM usage for the experiments amounted to approximately $5~\text{M}$ tokens.

\section*{Acknowledgments}
This work was made possible by the funding of the Federal Ministry of Research, Technology and Space (Bundesministerium für Forschung, Technologie und Raumfahrt) funding line ``KMU-Innovativ''.

We thank Tilo Wend and Lado Sirdadze of our collaborator Rulemapping Group for extensive discussions and valuable insights about legal formalizations.

\section*{AI Use Disclosure}
We used AI assistance throughout the research process to discuss and
evaluate approaches and design decisions, but made all final decisions regarding the method and design of the experiment ourselves. 
We used AI agents to implement the experiment code. At several stages we checked the
implemented features against the plan. We performed these checks partly manually, partly with AI assistance. Where we used AI to check code, the check ran in a separate session to ensure that the reviewing model evaluated the code against the plan without access to the context in which that code had been written.  
In Section~\ref{sec:related}, we used AI to identify relevant literature and to extract the relevant passages from cited papers.
In Section~\ref{sec:results}, we used AI to identify and discuss potentially interesting aspects and patterns in the results data, but made all final decisions ourselves.
In writing the paper, we used AI assistance to draft or revise individual
paragraphs within a pre-defined scope. We checked, edited, and approved
all generated text.
We reviewed and verified the accuracy of all results. We take full responsibility for this work.

\bibliography{custom}

\appendix

\section{Models}
\label{sec:appendix-models}

Table \ref{tab:models} shows the models used for creating formalizations of the EU provisions.

\begin{table}[h]
  \centering
  \small
  \begin{tabular}{@{}ll@{}}
    \toprule
    \textbf{Model} & \textbf{Provider} \\
    \midrule
    Claude Opus 4.6        & Anthropic \\
    GPT-5.4                & OpenAI \\
    Gemini 3.1 Pro         & Google \\
    DeepSeek-V3.2 (685B)   & DeepSeek \\
    Qwen 3.5 (397B)        & Alibaba \\
    Kimi K2.5 (1.1T)       & Moonshot \\
    Mistral Large 3 (675B) & Mistral \\
    MiniMax M2.7 (229B)    & MiniMax \\
    GLM-5.1 (754B)         & Zhipu AI \\
    \bottomrule
  \end{tabular}
  \caption{LLMs used for formalization ($N = 9$).}
  \label{tab:models}
\end{table}

\section{Worked Example}
\label{sec:appendix-worked-example}

This example follows the steps of Section~\ref{sec:methods} for
illustration.  The provision and all data are constructed manually
and are not drawn from the experimental data.

\subsection{Provision}

\begin{figure}[!t]
\begin{lawquote}[title={Data Retention}]
\small
1.~A controller may retain personal data where a legal basis for the
processing exists and no ground for deletion applies.
\smallskip\par
2.~The personal data has to be deleted if
\begin{enumerate}[label=(\alph*), nosep, topsep=3pt,
                  leftmargin=1.9em, labelsep=0.4em]
  \item the retention period laid down in the controller's retention
        schedule has expired;
  \item the data subject has requested erasure of the data.
\end{enumerate}
\end{lawquote}
\caption{An intentionally ambiguous, fictitious legal provision.}
\label{fig:running-norm}
\end{figure}

\begin{figure*}[!t]
  \centering
  \includegraphics[width=\textwidth]{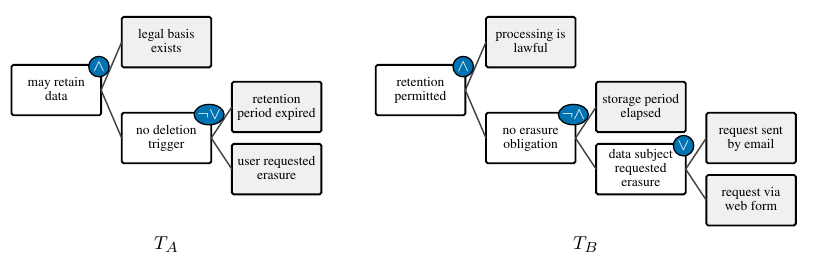}
  \caption{Two formalizations of the provision shown in Figure~\ref{fig:running-norm}.}
  \label{fig:running-example}
\end{figure*}

Figure~\ref{fig:running-norm} states the provision.  The construction
of paragraph~2 intentionally leaves open whether the two conditions
for a deletion have to be fulfilled alternatively or cumulatively.

\subsection{Formalization}

Figure~\ref{fig:running-example} shows two formalizations of the
provision, $T_A$ and $T_B$.  $T_A$ interprets the norm such that
either condition is sufficient to trigger the deletion, while $T_B$
requires both conditions to be fulfilled.  Additionally, $T_B$
resolves the erasure request into the channels through which it can
be made, where $T_A$ leaves it atomic.

\subsection{Matching}

\begin{figure*}[!t]
  \centering
  \includegraphics[width=\textwidth]{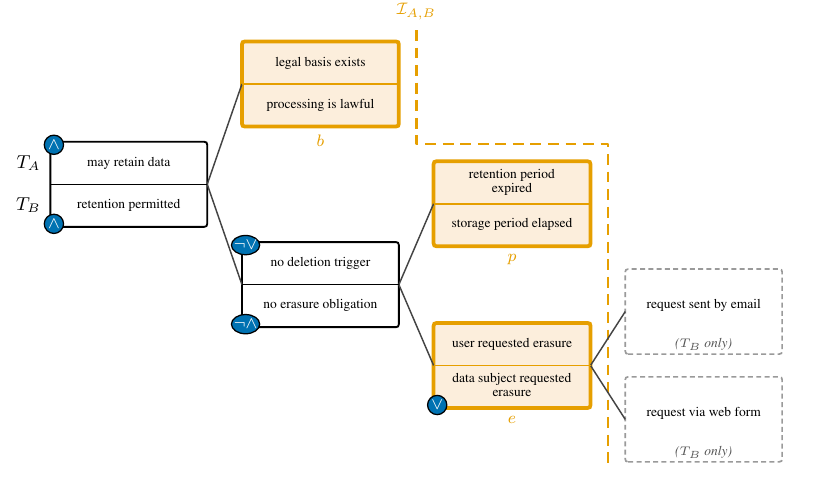}
  \caption{The two formalizations of
    Figure~\ref{fig:running-example} after matching.  The top half of
    each box is $T_A$, the bottom half $T_B$.  The interface
    $\mathcal{I}_{A,B}$ (orange) holds the deepest nodes still
    matched to each other and supplies the Boolean variables $b$, $p$
    and $e$; it sits at different depths on different branches.
    Everything to the right of it belongs to $T_B$ alone and is cut
    off.}
  \label{fig:interface-example}
\end{figure*}

Matching (Section~\ref{sec:methods-matching}) groups the nodes that
describe the same concept across the two formalizations; each such
group is an \emph{equivalence class} (EC).  Here the five concepts
the two formalizations share form one EC each, shown in
Figure~\ref{fig:interface-example}.  The two channel nodes of $T_B$
have no counterpart in $T_A$ and therefore stay unmatched.

\subsection{Interface}

The deepest matched nodes are $b$, the legal basis, $p$, the elapsed
period, and $e$, the erasure request.  They form the interface
$\mathcal{I}_{A,B} = \{b, p, e\}$
(Section~\ref{sec:methods-edge-case-analysis-interface-problem}).
Note that $e$ belongs to the interface although it is an internal
node of $T_B$. Its two child nodes, describing the valid request
channels, have no counterpart in $T_A$ and therefore cannot be
compared.  They are ignored in the subsequent steps and $e$ serves as
an atomic variable.

All five nodes of $T_A$ are on or above the interface, so
$\mathrm{cov}(T_A, \mathcal{I}_{A,B}) = 5/5 = 1$.  Of the seven nodes
of $T_B$, five are on or above, so
$\mathrm{cov}(T_B, \mathcal{I}_{A,B}) = 5/7 \approx 0.71$.  The
pairwise coverage is the geometric mean of the two,
$\mathrm{cov}(T_A, T_B) = \sqrt{5/7} \approx 0.85$.

\subsection{Edge Cases}

Reading the matched nodes above the interface as operators over the
interface variables (Section~\ref{sec:methods-edge-case-extraction})
gives
\begin{align*}
  f_A &= b \wedge \neg(p \vee e), \\
  f_B &= b \wedge \neg(p \wedge e),
\end{align*}
and hence $f_A \oplus f_B = b \wedge (p \oplus e)$.  The disagreement
function has exactly two prime implicants, $\{b, p, \neg e\}$ and
$\{b, \neg p, e\}$, and the cover returned by the procedure of
Appendix~\ref{sec:appendix-pi-enumeration} contains both.  The two
formalizations reach opposite conclusions whenever a legal basis
exists and exactly one of the two conditions of paragraph~2 is met.

Under either prime implicant, the deepest matched node whose two
sub-formulas are forced to differ is the deletion node, which is
therefore the root cause
(Section~\ref{sec:methods-edge-case-analysis-root-cause-identification}).

\subsection{Verbalization}

\begin{figure}[!htb]
\begin{casebox}[title={Verbalization of $\{b, p, \neg e\}$}]
\small
A hospital holds a patient's treatment records on a statutory legal
basis.  The retention period set out in its retention schedule
expired last month.  The patient has made no request for erasure.
\smallskip\par
\textit{May the hospital continue to retain the records?}
\smallskip\par
$T_A$: no.\quad $T_B$: yes.
\end{casebox}
\caption{A verbalization of one of the two edge cases.}
\label{fig:running-case}
\end{figure}

Figure~\ref{fig:running-case} shows an exemplary verbalization of one
of the two prime implicants
(Section~\ref{methods-verbalization}).  Applied to it, $T_A$
concludes that the records may no longer be retained and $T_B$ that
they may.

\FloatBarrier

\section{Edge Case Enumeration}
\label{sec:appendix-pi-enumeration}

We compute a non-redundant prime-implicant cover of the disagreement
function $\varphi := f_A \oplus f_B$ by SAT-iterative search.
The procedure maintains a growing set of \emph{blocking clauses}
$\beta$ that excludes already-covered sub-cubes, and uses a second
solver instance over $\neg\varphi$ to test the implicant property
needed to prime a candidate cube.
Figure~\ref{fig:pi-algorithm} states the procedure; the steps
below comment on each line.  Literals are dropped in a fixed order
(we sort the core by variable name), which makes the returned cover
deterministic.

\begin{figure}[t]
\small
\setlength{\tabcolsep}{0pt}
\renewcommand{\arraystretch}{1.15}
\begin{tabular}{@{}l@{}}
\toprule
\textbf{Input:} $\varphi := f_A \oplus f_B$ over $\mathcal{I}_{A,B}$ \\
\textbf{Output:} irredundant prime-implicant cover $\Pi$ \\
\midrule
$\Pi \gets \emptyset$, \quad $\beta \gets \top$ \\
\textbf{while} $\varphi \wedge \beta$ is \textsc{sat} \textbf{do} \\
\hspace{1.0em} $\mathbf{m} \gets$ model of $\varphi \wedge \beta$ \\
\hspace{1.0em} $C \gets \mathrm{unsatCore}(\neg\varphi, \mathbf{m})$ \\
\hspace{1.0em} $\pi \gets \mathbf{m}|_C$ \\
\hspace{1.0em} \textbf{for} $\ell \in C$ in fixed order \textbf{do} \\
\hspace{2.0em} \textbf{if} $\neg\varphi \wedge (\pi \setminus \ell)$
   is \textsc{unsat} \\
\hspace{3.0em} \textbf{then} $\pi \gets \pi \setminus \ell$ \\
\hspace{1.0em} $\Pi \gets \Pi \cup \{\pi\}$, \quad
   $\beta \gets \beta \wedge \neg\pi$ \\
\textbf{return} $\Pi$ \\
\bottomrule
\end{tabular}
\caption{Edge-case enumeration.  $\beta$ accumulates one blocking
  clause $\neg\pi$ per emitted prime implicant, so each iteration
  searches only the part of $\varphi$ no earlier implicant covers.
  $\mathbf{m}|_C$ is the minterm $\mathbf{m}$ restricted to the
  variables in the unsat core $C$.}
\label{fig:pi-algorithm}
\end{figure}

\begin{enumerate}
  \item \emph{Find an uncovered disagreement.}
        Check $\varphi \wedge \beta$, where $\beta$ is the
        conjunction of the blocking clauses accumulated so far
        (initially empty, i.e.\ $\top$).  If unsatisfiable, the
        cover is complete; return the accumulated cover.  Otherwise
        let $\mathbf{m}$ be a returned satisfying assignment ---
        i.e.\ a minterm of $\varphi$ that no previously emitted
        prime implicant covers.
  \item \emph{Shrink to an implicant via unsat core.}
        Treating the literals of $\mathbf{m}$ as Z3 assumptions,
        check $\neg\varphi \wedge \mathbf{m}$.  Since
        $\varphi(\mathbf{m}) = \texttt{true}$, this is unsatisfiable;
        the unsat core is a subset of the literals of $\mathbf{m}$
        that already forces $\varphi$ to hold under every completion
        of the remaining variables, hence an implicant of $\varphi$.
  \item \emph{Make the implicant prime by greedy literal drop.}
        Iterate over the literals of the core in a deterministic
        order; tentatively remove each one and verify (via a
        push/pop check on $\neg\varphi$ with the reduced cube as
        assumptions) that the remainder is still an implicant.
        Keep the literal otherwise.  The result $\pi$ is prime by
        construction.
  \item \emph{Block and iterate.}
        Add the clause $\neg\pi$ (i.e.\ the negation of the cube
        $\pi$) to $\beta$ and return to step~1.
\end{enumerate}

Each iteration emits exactly one prime implicant and removes at
least one previously uncovered minterm, so the loop terminates after
at most $|\varphi^{-1}(\texttt{true})|$ steps.  The resulting cover
is irredundant by construction --- every prime implicant eliminates
a sub-cube that no prior one covered --- but is not guaranteed to be
of minimum cardinality, since the literal-drop order in step~3 is
greedy rather than exhaustive.  In practice we observe per-pair
covers ranging from $1$ to $58{,}283$ prime implicants
(Section~\ref{sec:results-edge-case-analysis}).

\section{Case Study Provisions and Formalizations}
\label{sec:appendix-case-studies}

For the two verbalized edge cases shown in
Table~\ref{tab:verbalizations}, we reproduce the operative legal
text and a side-by-side diagram of the two disagreeing
formalizations.  The diagrams use the same visual language as
Figure~\ref{fig:formalization-concept}: rounded boxes for ECs,
operator badges for non-leaf nodes.  The disagreement path is
drawn left-to-right.  Matched ECs are split by a horizontal divider
into a top half (first formalization) and a bottom half (second),
each carrying that formalization's verbatim label and its operator
badge; unmatched nodes are single-sided and annotated below the
label with the contributing formalization.

\subsection{Art.~5 AI Act}
\label{sec:appendix-case-ai-act}

\begin{figure*}[!ht]
\begin{lawquote}[title={Art.~5 AI Act --- excerpt}]
1.~The following AI practices shall be prohibited: 

[\dots]

(g)~the
placing on the market, the putting into service for this specific
purpose, or the use of biometric categorisation systems that
categorise individually natural persons based on their biometric
data to deduce or infer their race, political opinions, trade union
membership, religious or philosophical beliefs, sex life or sexual
orientation; this prohibition does not cover any labelling or
filtering of lawfully acquired biometric datasets, such as images,
based on biometric data or categorizing of biometric data in the
area of law enforcement.

[\dots]
\end{lawquote}
\end{figure*}

\begin{figure*}[!ht]
  \centering
  \includegraphics[width=\textwidth]{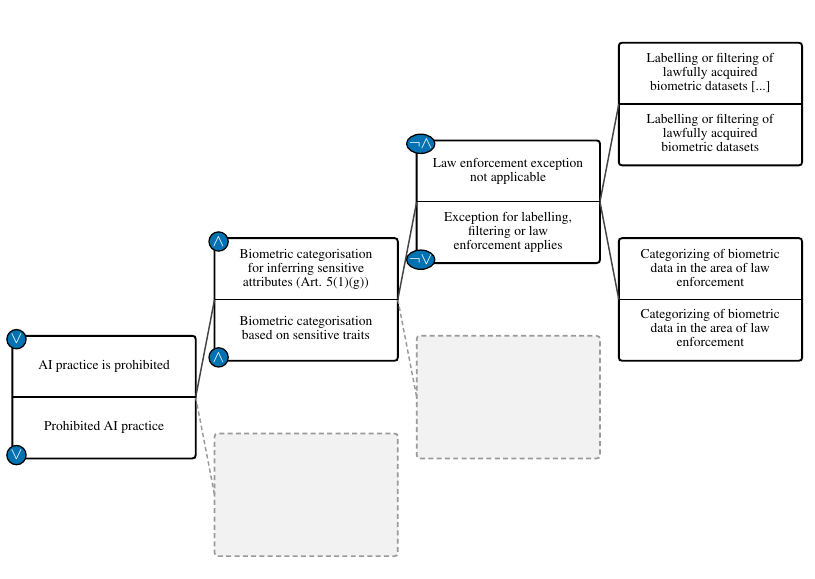}
  \caption{Matched formalizations of Art.~5(1)(g) AI~Act by
    Claude Opus~4.6 (top half of each node) and Gemini~3.1~Pro
    (bottom half).
    The two formalizations agree on the
    high-level structure but diverge on the operator at
    ``labelling/filtering exception not applicable'': Claude treats
    the carve-out as NAND of conditions, 
    Gemini as a NOR. 
    Dashed grey nodes indicate unshown parts of the formalizations.}
  \label{fig:case-ai-act}
\end{figure*}

\FloatBarrier
\subsection{Art.~3 UCTD}
\label{sec:appendix-case-uctd}

\begin{figure*}[!ht]
\begin{lawquote}[title={Art.~3 UCTD (Directive 93/13/EEC)}]
1.~A contractual term which has not been individually negotiated
shall be regarded as unfair if, contrary to the requirement of good
faith, it causes a significant imbalance in the parties' rights and
obligations arising under the contract, to the detriment of the
consumer.

2.~A term shall always be regarded as not individually negotiated
where it has been drafted in advance and the consumer has therefore
not been able to influence the substance of the term, particularly
in the context of a pre-formulated standard contract.

The fact that certain aspects of a term or one specific term have
been individually negotiated shall not exclude the application of
this Article to the rest of a contract if an overall assessment of
the contract indicates that it is nevertheless a pre-formulated
standard contract.

Where any seller or supplier claims that a standard term has been
individually negotiated, the burden of proof in this respect shall
be incumbent on him.

3.~The Annex shall contain an indicative and non-exhaustive list of
the terms which may be regarded as unfair.
\end{lawquote}
\end{figure*}

\begin{figure*}[!ht]
  \centering
  \includegraphics[width=\textwidth]{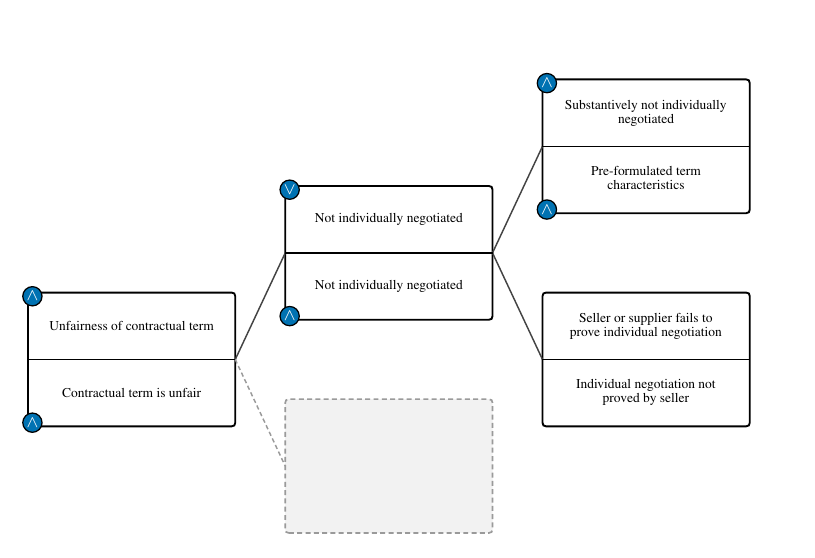}
  \caption{Matched formalizations of Art.~3 UCTD by GLM-5.1
    (top half of each node) and Kimi-K2.5 (bottom half).  The divergence
    sits at ``not individually negotiated'': GLM treats it as a
    disjunction (either pre-drafting or the burden-of-proof clause
    suffices), Kimi as a conjunction (both must hold).  
    The dashed grey node indicates unshown parts of the formalizations. For clarity
    the visualization was slightly simplified.}
  \label{fig:case-uctd}
\end{figure*}

\FloatBarrier
\section{LLM Prompts}
\label{sec:appendix-prompts}

Figures~\ref{fig:prompt-create-rulemap-1}--\ref{fig:prompt-verbalize-correction}
reproduce the prompt templates used in each stage of the pipeline,
verbatim from the codebase.  Placeholders enclosed in double curly
braces (e.g.\ \verb|{{SCHEMA}}|, \verb|{{LAW_TEXT}}|) are
substituted at runtime.

\begin{figure*}[!ht]
\begin{promptbox}[title={\texttt{create\_rulemap.md} \textnormal{(part 1/3)}}]
\begin{lstlisting}[style=prompt, firstline=1, lastline=32]
# Task

Create a Rulemap for the law text provided below.

Think of it as an assessment schema ("Prüfungsschema") a lawyer or judge would use:
1. You will be provided with the central legal consequence of the norm below
2. Build a decision tree a legal expert would use to determine whether that consequence applies in a specific case given the text of the law.

# Methodology

- **Do not** merely reframe the statutory wording. Use your knowledge of EU law and the relevant legal area to build the assessment schema as a domain expert would.
- If the law references **another law**, create a leaf node with the reference — do not resolve it.
- If the law references **another part of the same text** given to you, resolve that reference in the tree.
- The text of the law may contain parts that are not relevant for the central legal consequence provided to you. Ignore these parts of the law, but include *everything* that *is* relevant 

# Structural Rules

1. **No single-child nodes.** Every non-leaf node must have at least two children.
2. **Granularity.** 
  - Be as granular as necessary to assess individual cases, but not more.
  - Leaf nodes should be either 
    - facts that are objectively assessable given a specific case or
    - common and general legal concepts that apply to that norm but are not specific to it
  - Focus on the matter specific to the norm to be formalized. If common and general legal concepts are to be assessed, do not resolve them in detail
3. **Labels.** 
  - Use labels a legal domain expert would use as headings when assessing the law
  - Do not cite authors in the labels
  - Do not make comments in the label about reasons for your choice or any other meta-comments on the label
  - Do not put long enumerations in the label, but use a short wording that summarizes the set of cases precisely.
4. **IDs.** Assign sequential integer IDs starting at 1 (root = 1).
5. **Name.** The top-level `name` field should be a short identifier for the law (e.g. `"Art. 101 TFEU"` or `"GDPR Art. 6"`).
\end{lstlisting}
\end{promptbox}
\caption{Formalization stage (part 1/3): task description, methodology,
  and structural rules.}
\label{fig:prompt-create-rulemap-1}
\end{figure*}

\begin{figure*}[!ht]
\begin{promptbox}[title={\texttt{create\_rulemap.md} \textnormal{(part 2/3)}}]
\begin{lstlisting}[style=prompt, firstline=33, lastline=70]
## Commented examples
  1. Theft
    - Bad: 
      - Suspect took away movable object belonging to another person [COMMENT: these labels are mere rewordings of the law text and don't make the established legal concepts explicit.]
        - object was movable
        - object belongs to another person 
        - Suspect took away the object (as defined by Schneider) [COMMENT: Do not make any comments on your label choices.]
    - Good:
      - Movable object belonging to another
        - Movable object
        - Belonging to another
      - Taking  [COMMENT: "Taking" is the established legal category to be checked when assessing Theft, thus it has to be part of the structure]
        - Custody of another person [COMMENT: "Custody" is another concept in the context of Theft, thus it must appear here although it is not named explicitely in the text of the law.]
        - Breaking of custody
          - Ending of original custody
          - against the will
        - Esablishment of new custody
  2. Sales contract
    - Bad:
      - Parties closed a sales contract
        - Party 1 made an offer 
        - Party 2 accepted the offer
        - Offer and acceptance are about the sale of an object 
        - Offer and acceptance match each other (ambigious, but mostly accepted) [COMMENT: Do not comment on characteristics of label or why you decided to choose it.]
        - No invalidity due to lack of legal capacity, avoidance, immorality, ursury or violation of statutory prohibitions [COMMENT: Do not put long enumerations in labels.]
          - lack of legal capacity [COMMENT: Reasons for invalidity are common to most contracts and not specifically related to sales contracts. Resolving them here is beyond the scope of the formalization.]
            - age < 7 years
            - ...
          - usury
            - ...
          - ...
    - Good:
      - Sales contract
        - offer
        - acceptance
        - Contract main content: Sale
        - No invalidity
\end{lstlisting}
\end{promptbox}
\caption{Formalization stage (part 2/3): commented examples illustrating
  good and bad node labels and granularity.}
\label{fig:prompt-create-rulemap-2}
\end{figure*}

\begin{figure*}[!ht]
\begin{promptbox}[title={\texttt{create\_rulemap.md} \textnormal{(part 3/3)}}]
\begin{lstlisting}[style=prompt, firstline=71, lastline=101]
# Node Properties

- **`operator`** (required for every node):
  - `"and"` — all children must be satisfied.
  - `"or"` — at least one child must be satisfied.
  - `"virtual"` — **leaf nodes only**; indicates no logical operator (terminal condition).
- **`negation`** (`boolean`, default `false`): 
  - An operator can be logically negated by setting negation to 'true'. 
  - That is, nodes with `operator` = `and`/`or` and `negation` = `true` are interpreted as `nand` and `nor`, respectively.
  - If you decide to set `negation` = `false`, you may also omit this field.

**`operator`** and **`negation`** define the logical operator each non-leaf nodes represents. Remember that the final rule map will be used to assess actual legal cases. Therefore, make sure that the values you set for **`operator`** and **`negation`** 
align with your legal interpretation of the norm and the operators of the other nodes.

**`children`** is a required field. If a node is a leaf node, still add the field with an empty list \[\] as value.

# Output Format

- Return **ONLY valid JSON**. No markdown formatting, no preamble, no explanations.
- The JSON must comply with the following schema (do not include the schema meta-properties in your output):

{{SCHEMA}}

# Law Text

{{LAW_TEXT}}

# Central legal consequence

{{LEGAL_CONSEQUENCE}}
\end{lstlisting}
\end{promptbox}
\caption{Formalization stage (part 3/3): node properties, output format,
  and runtime placeholders.}
\label{fig:prompt-create-rulemap-3}
\end{figure*}

\begin{figure*}[!ht]
\begin{promptbox}[title={\texttt{create\_rulemap\_correction.md}}]
\begin{lstlisting}[style=prompt]
Your previous output failed validation. The error report is below.

## How to read the error report

The validator performs two checks:

1. **JSON Schema validation** — Your output must conform to the rulemap schema. Common schema errors include:
   - Missing required fields (`name`, `root`, `id`, `label`, `children`)
   - Wrong types (e.g., `id` must be an integer, `operator` must be one of `"and"`, `"or"`, `"virtual"`)
   - Invalid `operator` values or extra properties not allowed by the schema
   - Leaf nodes must have `children: []` (an empty array, not omitted)

2. **Unique ID check** — Every node `id` in the tree must be unique. If the same integer ID appears on multiple nodes, you will see: `Duplicate id detected in rulemap: <id>`.

If the error says "Expected a JSON object", your response was not valid JSON or was wrapped in markdown formatting. Return **only** the raw JSON object.

## Error report

{{ERROR}}

## Instructions

Fix the issues described above and return only the corrected, valid JSON. Do not include markdown code fences, explanations, or any text outside the JSON object.
\end{lstlisting}
\end{promptbox}
\caption{Formalization stage: follow-up prompt sent when the LLM's
  output fails JSON schema validation.}
\label{fig:prompt-create-rulemap-correction}
\end{figure*}

\begin{figure*}[!ht]
\begin{promptbox}[title={\texttt{matching.md}}]
\begin{lstlisting}[style=prompt]
__Input:__

1. __N Rule Maps (JSON):__ Each is a tree-structured formalization of the same legal norm, created independently. Nodes represent legal conditions connected by logical operators (`and`, `or`, `virtual`).
2. __Norm Text:__ The actual legal text to resolve ambiguities.

__Task:__ Group the nodes from all N rule maps into __equivalence classes__ (ECs). Each EC groups nodes across rule maps that represent the same legal concept. Nodes without counterparts in other rule maps form singleton ECs.

__Matching Strategy:__

Match nodes based on their conceptual legal role in the respective trees. Refer to the text of the law to identify the legal meaning of the nodes in the context of the whole structure of each tree.
If a node has no counterpart in any other tree (e.g., because of differing granularity or entirely different structuring), place it in a __singleton EC__ (an EC with only one member).

__Mandatory Rules:__

1. __At most one node per rule map per EC:__ Each EC may contain at most one node from each rule map. This means no two nodes from the same rule map can be in the same EC.

2. __Exhaustiveness:__ Every node from every rule map must appear in exactly one EC. No node may be omitted or appear in multiple ECs.

3. __Roots matched:__ The root nodes of all rule maps must be in the same EC.

4. __EC Labels:__ Each EC must have a descriptive `label` field that summarizes the legal concept it represents (e.g., "Applicability of Article 101 TFEU", "undertaking or association of undertakings", "agreement between undertakings"). This label should reflect the common legal meaning, not just copy a single node's label.

5. __No cycles:__ The parent-child relationships derived from the tree structures must form a directed acyclic graph (DAG) when projected onto the ECs. Concretely: if node A is an ancestor of node B in one tree, then the EC containing A must not be a descendant of the EC containing B in any other tree's structure.

6. __Schema Compliance:__

   - Your result __must__ conform to the following JSON schema:
   `{{SCHEMA}}`

7. __No Hallucinations:__ You must use the exact node objects (ID, Label, Operator) provided in the input. Do not invent new IDs or labels. Copy label and operator strings exactly as given. For `rulemap_id`: each rule map is presented with a header `=== File N: filename.json ===`. The `rulemap_id` is the **last underscore-separated segment** of the filename (without `.json`). Examples: `9_gdpr_anthropic_claude-opus-4-6.json` → `claude-opus-4-6`; `9_gdpr_ollama_minimax-m2.7.json` → `minimax-m2.7`.

8. __Output Format:__ Return __ONLY valid JSON__. No markdown formatting (like ```json), no preamble, no explanations.

# Law Text

{{LAW_TEXT}}

## Rule Maps

{{RULEMAPS}}
\end{lstlisting}
\end{promptbox}
\caption{Matching stage: prompt for $N$-way matching of nodes across
  rule maps.}
\label{fig:prompt-matching}
\end{figure*}

\begin{figure*}[!ht]
\begin{promptbox}[title={\texttt{matching\_correction.md}}]
\begin{lstlisting}[style=prompt]
Your previous output failed validation. The error report is below.

## How to read the error report

The validator performs a JSON Schema check followed by 7 integrity checks:

1. **JSON Schema validation** — The output must match the N-way matching schema. Common errors: missing `label` or `members` in an equivalence class, empty `members` array, wrong field types.

2. **Rulemap ID uniqueness** — Each `rulemap_id` referenced in members must match one of the input rule map filenames. If you see "invalid rulemap ref", you used a filename that doesn't exist in the input.

3. **One node per rulemap per EC** — Each rule map may contribute at most one node to each equivalence class. If you see a "collision" error, the same `rulemap_id` appears twice in one EC's members.

4. **Node coverage** — Every node from every input rule map must appear in exactly one equivalence class.
   - "missing nodes" = nodes from the input that you didn't assign to any EC.
   - "hallucinated nodes" = node IDs you referenced that don't exist in the input rule maps.
   - "duplicate nodes" = the same node appears in multiple ECs.

5. **Label consistency** — The `label` and `operator` you report for each member must match the actual values in the source rule map. Mismatches indicate you copied labels incorrectly.

6. **Roots in same EC** — All root nodes (the top-level node of each rule map) must be grouped into the **same** equivalence class.

7. **DAG check** — The parent-child relationships implied by the ECs must form a directed acyclic graph. If you see a "cycle" error, the grouping creates circular dependencies.

## Error report

{{ERROR}}

## Instructions

Fix the issues described above and return only the corrected, valid JSON. Do not include markdown code fences, explanations, or any text outside the JSON object.
\end{lstlisting}
\end{promptbox}
\caption{Matching stage: follow-up prompt sent when the matching
  output fails validation.}
\label{fig:prompt-matching-correction}
\end{figure*}

\begin{figure*}[!ht]
\begin{promptbox}[title={\texttt{verbalize\_edge\_case.md} \textnormal{(part 1/3)}}]
\begin{lstlisting}[style=prompt, firstline=1, lastline=40]
# Task: Verbalize a Formal Edge Case into a Concrete Legal Scenario

Multiple LLM-generated rule maps for the same EU legal provision disagree on
whether the provision **applies** under certain input conditions. SAT analysis
has identified a *minimal* fact pattern (the **prime implicant**) that forces
the disagreement: any scenario matching these facts produces the disagreement,
regardless of what other facts are true.

Your job is to write a short, realistic scenario illustrating exactly this fact
pattern, pose the applicability question it provokes, and explain how the two
rule maps disagree as a question of *legal interpretation* — so a human reader
can judge which interpretation is more faithful to the provision.

## Inputs

You are given the raw formal data below. No prose rendering has been applied;
you must read the JSON objects and do all narrative work yourself.

### 1. Legal provision

{{PROVISION_TEXT}}

### 2. Projected EC graph (two rule maps)

A directed graph of *equivalence classes* (ECs) restricted to the two rule
maps in this disagreement. Every EC has an `id`, a `canonical_label`, and a
`members` list — one entry per rule map that contains this EC, recording the
logical `operator` (`and` / `or` / `virtual`) and `negation` at that node.
Edges carry `supporting_rulemaps`: which of the two rule maps uses that
parent → child link.

A rule map's tree for this provision is reconstructed by walking the edges
filtered on its `rulemap_id` and reading the operator off each EC's
corresponding member entry. A `virtual` operator marks a leaf (atomic
condition).

```json
{{PROJECTED_EC_GRAPH}}
```
\end{lstlisting}
\end{promptbox}
\caption{Verbalization stage (part 1/3): task description and inputs
  (legal provision, projected EC graph).}
\label{fig:prompt-verbalize-1}
\end{figure*}

\begin{figure*}[!ht]
\begin{promptbox}[title={\texttt{verbalize\_edge\_case.md} \textnormal{(part 2/3)}}]
\begin{lstlisting}[style=prompt, firstline=41, lastline=66]
### 3. Stipulated facts (prime implicant)

Each entry is an EC-graph node augmented with `"value": true|false`. These
assignments **must hold** in the scenario you construct. Together they force
the two rule maps to disagree; no other facts are needed.

```json
{{STIPULATED_PI}}
```

### 4. Sides of the disagreement

- Rule map concluding the provision **applies**: `{{RM_TRUE}}`
- Rule map concluding the provision **does not apply**: `{{RM_FALSE}}`

### 5. Deepest forced-divergence points (root causes)

Raw symbolic records for the ECs where the structural divergence is deepest
(one entry per root cause). Each dict carries `ec_id`, `ec_label`, and
per-side operator / negation / children information. Use these to locate
the disagreement inside the projected graph.

```json
{{ROOT_CAUSES}}
```
\end{lstlisting}
\end{promptbox}
\caption{Verbalization stage (part 2/3): inputs continued (stipulated
  facts, sides of the disagreement, deepest forced-divergence points).}
\label{fig:prompt-verbalize-2}
\end{figure*}

\begin{figure*}[!ht]
\begin{promptbox}[title={\texttt{verbalize\_edge\_case.md} \textnormal{(part 3/3)}}]
\begin{lstlisting}[style=prompt, firstline=67, lastline=134]
## Instructions

Produce a single JSON object with the following fields.

### `title`

A short (<= 10 words) descriptive title for the scenario.

### `scenario`

A factual narrative (2-6 sentences) describing a plausible real-world
situation in which every stipulated fact holds with the indicated
`true` / `false` value. Hard rules:

- **Facts only.** State what the actors did and what the factual
  circumstances are. Do *not* state, imply, or evaluate legal conclusions
  (e.g. do not write "the clause is unfair", "the processing is unlawful",
  "the exception applies"). Whether the provision applies is exactly the
  contested question, it must not be pre-judged by the narrative.
- **No parentheticals.** Do not insert inline citations or clarifications in
  parentheses or brackets such as "(Art. 6(1)(f) GDPR)", "(legitimate
  interest)", "[i.e. ...]". Legal citations belong in the legal provision
  text, not in the scenario. If a factual label is needed, integrate it as
  plain prose.
- **Concrete.** Use realistic parties, data, and circumstances. Avoid
  abstract placeholders.
- **Minimal.** Omit narrative detail that is not required to establish a
  stipulated fact.
- **Rely on plausibility for the default case.** Do not state a stipulated
  fact explicitly when its value is what any reasonable reader would already
  assume from the other facts in the scenario.  Example: if the scenario
  describes an eight-year-old child, you do *not* need to write
  "the child is unmarried" to establish a stipulated ``married = false``.
  Unmarried is the default the reader will infer, and stating it is awkward.
  State the fact explicitly *only* when its stipulated value contradicts
  the default expectation.

### `question`

One short neutral question (<= 25 words) asking whether the provision applies
to this scenario. Do not hint at an answer.

### `divergence_analysis`

A concise legal-interpretation contrast between the two sides. Explain, in
legal terms, *what interpretive question* divides the two formalizations on
this fact pattern and *why each reading is defensible* given the provision
text. Focus on the substantive legal disagreement — the contested scope,
condition, or exception — not the formal structure.

Do not describe the rule-map formalism, operators, or EC-graph structure.
Do not use "AND", "OR", "NOT-OR", "negation", "EC IDs", "virtual nodes",
"prime implicants", "equivalence classes", or similar technical vocabulary.
A reader who has never seen a rule map must be able to follow it.

### `interpretation_type`

Classify the disagreement as one of:
- `"legal_ambiguity"` — the provision text is genuinely open to both
  readings; a reasonable legal expert could defend either.
- `"formalization_error"` — one reading appears to misparse the
  provision (e.g. treats disjunctive conditions as conjunctive, misreads
  scope of an exception, omits a required element).

## Output format

Return **only** the JSON object, with no markdown fences, no preamble, and no
trailing commentary.
\end{lstlisting}
\end{promptbox}
\caption{Verbalization stage (part 3/3): instructions for each output
  field, and output format.}
\label{fig:prompt-verbalize-3}
\end{figure*}

\begin{figure*}[!ht]
\begin{promptbox}[title={\texttt{verbalize\_correction.md}}]
\begin{lstlisting}[style=prompt]
Your previous response failed validation.  The validator reports:

{{ERROR}}

## Instructions

Fix the issues above and return a corrected JSON object that conforms to
the schema.  Do not include markdown code fences, explanations, or any
text outside the JSON object itself.
\end{lstlisting}
\end{promptbox}
\caption{Verbalization stage: follow-up prompt sent when the
  verbalization output fails validation.}
\label{fig:prompt-verbalize-correction}
\end{figure*}

\end{document}

%% file: sections/introduction.tex
\section{Introduction}
\label{sec:introduction}

\begin{figure*}[!t]
  \centering
  \includegraphics[width=\textwidth]{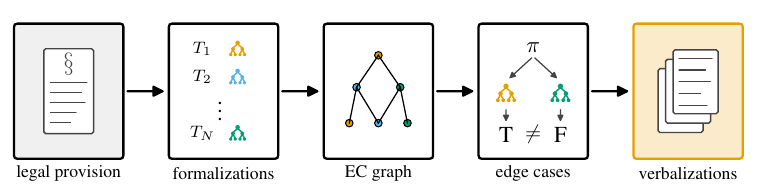}
  \caption{Analysis pipeline. From a legal provision, $N$ LLMs each
    produce a formalization; an LLM matches nodes across the
    formalizations into an equivalence-class (EC) graph; pairwise SAT
    analysis enumerates edge cases; an LLM verbalizes them for legal review.}
  \label{fig:pipeline}
\end{figure*}

Concerns about the ambiguity of natural-language legal text and calls
to represent the law in a formalized, unambiguous manner go back at
least to the early days of symbolic logic \cite{allen1957symbolic}.
The pay-off, if achievable, would be substantial: a formal
representation makes the law machine-accessible and enables automated
legal decisions in a broad range of settings. The idea has regained
traction with the rise of LLMs, which possess enough contextual
understanding to --- at least in principle --- produce such
formalizations directly from statutory text
\cite{janatian2023text,graus2026legal}.

However, the call for formalization is to be taken with caution. Any
formalization makes implicit interpretive choices, and foreseeing the
full consequences of those choices is hard: a sound-looking
formalization can yield an unjust verdict in a future case its
drafter did not anticipate. The problem is even more pressing when
the formalization is produced by an LLM, whose interpretive choices
are opaque to the reviewer.

We therefore address the question of how to make different
formalizations of the same legal provision systematically comparable
with respect to their outputs. We present a method that first
matches multiple formalizations of a provision against one another,
then uses SAT-based analysis to enumerate every input assignment on
which any two of them disagree. We apply the method to
nine LLM-generated formalizations of ten EU provisions across the
GDPR, the AI Act, the DMA, and other instruments. 
We find that outcome disagreement persists even between
formalizations whose matching aligns most of their structure, so
structural comparison alone does not reveal it.
We perform a case study and show that verbalization helps to surface
deeply rooted divergences. For example, we find that the models split evenly into
two camps on whether an AI system performing biometric categorization 
by race for law enforcement would be prohibited according to Article~5 AI Act, 
an interpretive divide that is also visible in the legal commentary on the provision. 

Specifically, we make the following contributions:

(i) a framework for comparing different formalizations
of the same legal provision by cases on which they disagree;

(ii) a verbalization method that turns SAT-derived edge cases
 into concrete factual scenarios for human review;

(iii) an empirical study of nine frontier LLMs on ten EU provisions, 
revealing systematic divergences that purely structural comparison cannot detect.\footnote{Experiment code and data are published at \url{https://doi.org/10.5281/zenodo.22128911}.}

%% file: sections/related_work.tex
\section{Related Work}
\label{sec:related}

\subsection{Legal Formalizations}
\label{sec:related-formalizations}

The approach to represent laws and legal provisions in formal algorithmic ways is not new. Notable early contributions include normalized legal drafting \cite{allen1978normalized} and the direct transcription to Prolog \cite{sergot1986british}. More recently, several approaches for dedicated legal programming languages appeared, such as Catala \cite{merigoux2021catala} for computational law, Stipula \cite{crafa2023stipula} for legal contracts, and L4 \cite{governatori2023defeasible} with defeasible legal-reasoning semantics; \citet{ma2021legislative} discuss the syntax such a representation should take.

While the process of formalizing the law was originally human-driven, the rise of LLMs and their fast adoption for general coding tasks quickly raised the question of whether they can applied to legal formalization as well. Notable work here include \citet{janatian2023text}, who explored the potential of LLMs for legal formalization and \citet{dal2025lost}, who used LLMs to represent EU regulation in knowledge graphs.

\citet{nguyen2026gdpr} generate GDPR
formalizations as condition--exception rule trees with LLM agents and
validate them through a panel of verifier agents and human review,
reporting that failures stem mostly from abstraction choices rather
than logical errors. While their method validates a single formalization
against expert judgment, ours compares several formalizations to
each other and needs no reference encoding.  
The authors also map a natural-language scenario on the generated formalizations,
however do not generate edge-case scenarios from logical differences.

\subsection{Matching}
\label{sec:related-matching}
The identification of corresponding concepts across heterogeneous
structured representations is studied as \emph{ontology matching},
with nearly two decades of benchmarking through the Ontology Alignment
Evaluation Initiative \cite{euzenat2011ontology}. Matchers exploit the
signals a curated ontology supplies, namely the labels of its entities,
the structure of its class and property hierarchy, the instances
populating its classes, and the axioms a reasoner can act on
\cite{otero_2015_ontology}. Established systems include LogMap
\cite{jimenez2011logmap} and AgreementMakerLight
\cite{faria2013agreementmakerlight}.

With the rise of LLMs, prompt-based matchers have entered the field.
\citet{macilenti2024prompting} report the limits of prompting alone,
finding that a prompting recipe carried from the OAEI benchmarks to a larger,
unpublished alignment loses most of its accuracy and fails on hierarchical
correspondences, where one concept is broader or narrower than the
other. They attribute the drop to the size of the candidate space and to the presence of such correspondences.
\citet{parciak_2024_schema} match schema elements from names and short
descriptions alone, in a setting where instances are
unavailable and labels are too cryptic for thesauri. Prompting one element
against a set of candidates outperforms a string-similarity baseline on
nearly every dataset they test, while prompting isolated pairs or
entire cross-products does worse.

Our setting falls inside these limits. Fewer than a thousand nodes
enter the matching of any single provision, and only equivalence
relations are required. Additionally, an LLM can be provided with unstructured context for the matching. In our case, we want to supply the text of the provision to improve interpretation of the formalization. We therefore decide to use an LLM for the
matching step.

\subsection{Equivalence Checking}
\label{sec:related-logic}
Checking whether two Boolean functions agree on every input is a
well-studied problem. Prime implicants \cite{quine1952simplifying}
are a compact way to summarize the cases in which two such functions
disagree. Their main application has been in hardware design, where
they are central to minimizing switching circuits and to checking
whether two alternative chip designs implement the same function
\cite{kuehlmann2002robust}. Today, equivalence checking is largely
handled by SAT and SMT solvers, descending from the DPLL procedure
\cite{davis1962machine} and available in modern tools such as
Z3 \cite{demoura2008z3}.

Outside the legal domain, \citet{li2024autoformalize} compare
multiple LLM formalizations of the same mathematical statement by
symbolic equivalence, using agreement among the candidates as a
selection signal.  We share the premise that equivalence between
independently generated formalizations is informative where no
reference exists, but use it to localize disagreement rather
than to select a winner. Additionally, unless some form of formal semantics and closed domain vocabulary is defined ahead of time, formalizations of realistically complex codified law usually have neither a theorem
prover to appeal to nor a shared vocabulary to begin with, which is what the matching and interface steps supply.

Logical analysis of legal artifacts has appeared only sporadically.
\citet{allen1957symbolic} uses a manually-crafted truth table to prove
that a formalized version of a part of the U.S.\ Internal Revenue
Code is equivalent to the actual law.
\citet{governatori2018deontic} use defeasible deontic logic to formally
represent a legal normative system and detect conflicts within it.
\citet{khoja2025contractcheck} formalize contract clauses in decidable
fragments of first-order logic and use SMT to detect internal
inconsistencies.
Adjacent work in business-process
compliance analyses how regulatory changes affect formal process
models~\cite{barrientos2026impact}.

Most recently, \citet{graus2026legal} generated LLM-based decision
models and evaluated them against a human-crafted gold standard
via outcome similarity, that is, the share of input combinations on
which model and gold standard reach the same result.  Our setup
differs on three counts.  First, we compare formalizations
against one another rather than against a gold standard, and so
make no assumption about correctness.  
Second, we find the input nodes for SAT analysis for each pair based on matching, rather than requiring a fixed set of inputs.  Third, we surface specific edge cases as concrete factual scenarios (Section~\ref{methods-verbalization}), rather than collapsing the comparison to an aggregate similarity score.

\subsection{Verbalization}
\label{sec:related-verbalization}
Making formal-verification artifacts accessible to non-experts is a
recognized concern in adjacent fields. The literature on
\emph{counterexample explanation} surveyed by
\cite{kaleeswaran2022slr} spans visual, trace-based, and
template-driven approaches to translating model-checking
counterexamples into more accessible forms.
\citet{moreira2025genai} suggest going further and use LLMs to improve template-based counterexample explanations.

A separate line of work uses LLMs to \emph{synthesize} concrete test
scenarios from higher-level specifications in safety-critical
domains, such as LeGEND \cite{tang2024legend} for autonomous-driving
systems and \cite{alazzoni2025smartcontract} for smart contracts. However, these works use test scenarios as inputs for system tests, instead of analyzing them as outputs.

%% file: sections/data_set.tex
\section{Dataset}
\label{sec:data}

We study 10 legal provisions selected from EU law (Table~\ref{tab:provisions}).
They span four domains: data protection (General Data Protection
Regulation, GDPR), consumer law (Unfair Contract Terms Directive,
UCTD; Unfair Commercial Practices Directive, UCPD), digital-market
and AI regulation (Digital Markets Act, DMA; AI Act), and public
procurement and state aid (Public Procurement Directive; Treaty on
the Functioning of the European Union, TFEU).
Provisions were selected according to three criteria:

(i) Complexity: the analysis of formalizations is only interesting
for norms with potentially differing or erroneous interpretations. We therefore chose provisions with multiple nested conditions and non-trivial interactions between them. This forces the LLM to make actual decisions about interpretation.

(ii) Self-containment: the provision should not refer extensively to 
other norms and should be mostly interpretable without having complete knowledge 
of the larger corpus of law.

(iii) Binary legal consequence: The provision must encode the decision of a clear, binary legal consequence to ensure that the root of every formalization is interpretable as a Boolean entitlement decision.

\begin{table}[t]
  \centering
  \small
  \setlength{\tabcolsep}{4pt}
  \begin{tabular}{@{}r@{\ }c@{\enspace}p{0.33\columnwidth}p{0.38\columnwidth}@{}}
    \toprule
    \# & & \textbf{Provision} & \textbf{Legal consequence} \\
    \midrule
    1  & \provswatch{1}  & Art.~6 GDPR   & Data processing is lawful \\
    2  & \provswatch{2}  & Art.~9 GDPR   & Processing of special-category data is prohibited \\
    3  & \provswatch{3}  & Art.~17 GDPR  & Data subject has right to erasure \\
    4  & \provswatch{4}  & Art.~22 GDPR  & Data subject has right not to be subject to automated decision-making \\
    \midrule
    5  & \provswatch{5}  & Art.~3 UCTD   & Contractual term is unfair \\
    6  & \provswatch{6}  & Art.~6--7 UCPD & Commercial practice is misleading \\
    \midrule
    7  & \provswatch{7}  & Art.~3 DMA    & Undertaking is designated as gatekeeper \\
    8  & \provswatch{8}  & Art.~5 AI Act & AI practice is prohibited \\
    \midrule
    9  & \provswatch{9}  & Art.~57 Proc.\ Dir. & Economic operator is excluded from procedure \\
    10 & \provswatch{10} & Art.~107 TFEU & State aid is incompatible with the internal market \\
    \bottomrule
  \end{tabular}
  \caption{Legal provisions and their central legal consequences.
           Numbers and colors identify provisions across all figures.}
  \label{tab:provisions}
\end{table}

%% file: sections/methods.tex
\section{Methods}
\label{sec:methods}

Figure~\ref{fig:pipeline} gives an overview of the full analysis pipeline. We provide a worked example in Appendix~\ref{sec:appendix-worked-example} to illustrate the process.

\subsection{Formalization}
\label{sec:methods-formalization}

\begin{figure}[t]
  \centering
  \includegraphics[width=\columnwidth]{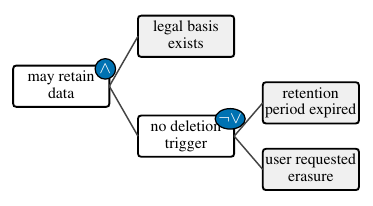}
  \caption{Illustration of a formalization. Internal nodes carry a 
    logical operator and the root represents the legal
    consequence to be decided on.}
  \label{fig:formalization-concept}
\end{figure}

A \emph{formalization} of a legal provision is a rooted, labeled tree
$T = (V, E)$.  Each node $v \in V$ carries a natural-language label
$\ell(v)$, and every non-leaf node additionally carries a logical operator
$\mathrm{op}(v) \in \{\textsc{and}, \textsc{or}, \textsc{nand},
\textsc{nor}\}$.
A formalization does not attempt to mirror the syntactic structure of the
provision text.  Instead, it captures the decision schema a legal expert
would follow when assessing whether the provision applies to a concrete
case, thereby making that reasoning structure explicit and machine-readable.

For our study we instruct $N = 9$ different LLMs to produce a
formalization of each provision; the full model list is given in
Appendix~\ref{sec:appendix-models} (Table~\ref{tab:models}). This
yields a set of trees $\mathcal{T} = \{T_1, \ldots, T_N\}$ per
provision.

\subsection{Matching}
\label{sec:methods-matching}

To compare formalizations of the same provision we need to identify
nodes across different trees that represent the same legal concept.
We call this step \emph{matching}.

Given $N$ formalizations $\mathcal{T} = \{T_1, \ldots, T_N\}$ for a
single provision, a matching groups all nodes across the $N$ trees
into \emph{equivalence classes} (ECs); we write $\mathcal{E}$ for the
full set of ECs.
Each EC $e \in \mathcal{E}$ is a set of nodes drawn from distinct
formalizations, that is, every formalization contributes at most one
node to~$e$. We interpret an EC as a legal concept present in the
participating formalizations.

\subsubsection{Choosing a Matching Method}
\label{sec:methods-matching-method}

Concept matching is the most complex step of the experiment.
A correct matching requires understanding (1)~the provision,
(2)~each formalization in its context, and (3)~individual nodes
in the context of their surrounding tree structure.
Tree-edit-distance metrics and embedding-similarity baselines we
tried in pilot experiments handle none of these requirements, and
classical ontology matching systems are designed for settings that
do not apply here (Section~\ref{sec:related-matching}). Following
\citet{parciak_2024_schema}, we therefore use an LLM for the
matching step.

Direct accuracy measurement is, however, infeasible: no benchmark
exists for matching formalizations of legal norms, and creating one
would itself require resolving the very ambiguity we study. Two
legal experts asked to match the same nodes are prone to disagree,
since interpretation of the formalization, like interpretation of
the law itself, is a matter of judgment. What we can assess is
reliability: if the same model produces consistent results across
independent runs, we have reason to trust it captures structural relationships rather than making arbitrary choices.  We
therefore conduct a consistency study across candidate models and
select the most consistent one for all subsequent matching runs
(Section~\ref{sec:results-matching-consistency}).

Consistency, however, does not exclude systematic error. A matcher might consistently fail to match related legal concepts or match unrelated ones. We therefore use the consistency study only to select among candidate models and treat matching correctness as an open limitation.

\subsection{Edge Case Analysis}
\label{sec:methods-edge-case-analysis}

Each formalization can be read as a Boolean function. The leaf nodes are input variables, internal nodes apply their operator to their children, and the root yields the legal consequence.

\subsubsection{Edge Cases}
\label{sec:methods-edge-case-analysis-prime-implicants}
Given two formalizations $T_A$, $T_B$ of the same legal provision over
a shared set of Boolean input variables $\mathbf{x} = (x_1, \ldots,
x_n)$, let $f_A(\mathbf{x})$ and $f_B(\mathbf{x})$ denote their
respective root functions. We characterize
edge cases through \emph{prime implicants}~\cite{quine1952simplifying}
of the disagreement function $f_A \oplus f_B$.  A prime implicant is
a partial assignment $\pi: S \to \{\texttt{true}, \texttt{false}\}$,
$S \subseteq \{x_1, \ldots, x_n\}$, such that (i) every completion of
the unassigned variables yields $f_A \oplus f_B = \texttt{true}$, and
(ii) no proper subset of $S$ satisfies (i). Each prime implicant captures a distinct, minimal pattern of
disagreement between $T_A$ and $T_B$.
We use \emph{prime implicant} and \emph{edge case} interchangeably throughout.

\subsubsection{The Interface Problem}
\label{sec:methods-edge-case-analysis-interface-problem}

A Boolean comparison requires a shared input vocabulary, but
different formalizations rarely share the same leaf nodes: one may
split a concept into three sub-conditions where another uses two.
The matching step's EC graph supplies this vocabulary.

We define the \emph{interface} $\mathcal{I}_{A,B} \subseteq
\mathcal{E}$ as the set of deepest ECs to which both $T_A$ and $T_B$
contribute a node, and write $\mathcal{E}^{\geq\mathcal{I}}$ for the
ECs lying on or above it: those in $\mathcal{I}_{A,B}$ together with
those on a path from a root to a member of $\mathcal{I}_{A,B}$.
Reading every EC above the interface as its operator applied to its
children, and every interface EC as an atomic variable, gives a
Boolean formula $f_A$ (resp.\ $f_B$) over the interface variables.
This is the analogue of cut-point equivalence checking in hardware
verification~\cite{kuehlmann2002robust}.
Appendix~\ref{sec:appendix-worked-example} works the construction
through on a small example.

The \emph{coverage} of $T_x$ measures how much of the formalization
lies on or above the interface:
\begin{equation}
  \mathrm{cov}(T_x, \mathcal{I}_{A,B})
    = \frac{|\mathcal{E}^x \cap \mathcal{E}^{\geq\mathcal{I}}|}
           {|\mathcal{E}^x|},
  \label{eq:coverage}
\end{equation}
with $\mathcal{E}^x$ the set of ECs to which $T_x$ contributes a
node.  For a pair we report the geometric mean of the two,
\begin{equation}
  \mathrm{cov}(T_A, T_B)
    = \sqrt{\mathrm{cov}(T_A, \mathcal{I}_{A,B}) \cdot
            \mathrm{cov}(T_B, \mathcal{I}_{A,B})}.
  \label{eq:coverage-pair}
\end{equation}
Low coverage means most of either tree is invisible to the Boolean
comparison, so we restrict to $\mathrm{cov}(T_A, T_B) \geq 0.4$
throughout, justified empirically in
Figure~\ref{fig:model_edginess} (left panel).

The interface is defined pairwise; a joint interface across all
$N$ formalizations would be pulled up by any structural outlier and
collapse to a few high-level concepts.

Two formalizations may encode the same concept with very different
granularity
(Appendix~\ref{sec:appendix-worked-example}).  The interface locates the deepest level at which they
still share a vocabulary, so whatever lies below it is cut off from
the Boolean comparison.  This is inherent to an outcome-based
analysis: where the decompositions part, no shared vocabulary is
left to compare over. We want to keep the method applicable to formalizations of
any structure. We therefore impose no granularity requirement. To make explicit how informative a comparison is, we report coverage, that is, the fraction of each
formalization the comparison takes in.

\subsubsection{Edge Case Extraction}
\label{sec:methods-edge-case-extraction}

For each pair, we build $f_A$ and $f_B$ as Boolean formulas over the
interface variables by recursively translating each formalization's
operators.  Edge cases are then extracted via a SAT-iterative search
using Z3~\cite{demoura2008z3} until every input assignment on which
$f_A$ and $f_B$ disagree is captured.
Appendix~\ref{sec:appendix-pi-enumeration} gives the procedure in detail.

\subsubsection{Root Cause Identification}
\label{sec:methods-edge-case-analysis-root-cause-identification}

A prime implicant identifies \emph{which} input conditions force
disagreement, but not \emph{where} in the formalizations it
originates.  Each EC above the interface induces one sub-formula per
formalization: the part of that formalization rooted at it.  A
\emph{root cause} of a prime implicant $\pi$ is an EC whose two
sub-formulas are forced to differ under every completion of $\pi$,
while no child of it is, in either formalization.  It is the deepest
point at which the two formalizations are already committed to
different answers.

\subsection{Verbalization}
\label{methods-verbalization}

Long lists of prime implicants are hardly interpretable by human reviewers.
We therefore \emph{verbalize} selected edge cases
into concrete factual scenarios in natural language 
that a legal expert can evaluate. 
Verbalizing every edge case is, however, both computationally infeasible and legally redundant: 
most edge cases within a pair are minor variants of a few underlying
disagreement patterns and equivalent constellations arise in multiple pairs.
We therefore apply a selection pipeline that reduces the raw edge-case sets 
to a small set of \emph{representative edge cases} per provision.

First, we discard pairs below the coverage
threshold introduced in
Section~\ref{sec:methods-edge-case-analysis-interface-problem}.
We further drop pairs whose root operators differ between $T_A$ and
$T_B$, since such pairs disagree at the topmost level of the
formalization and offer no fine-grained edge case worth verbalizing.

Second, for each surviving pair, we group the edge cases
by their root causes
(Section~\ref{sec:methods-edge-case-analysis-root-cause-identification}).
Each group has a canonical \emph{signature}: at every deepest root
cause, the operator, negation, and child set on the side that
concludes \texttt{true} and on the side that concludes \texttt{false}.
Edge cases with the same signature share both where the disagreement
originates and how the formalizations diverge there. They therefore
describe the same kind of legal disagreement.

Third, within each signature class, we
select the edge case with the fewest fixed variables as the \emph{representative}: 
a shorter edge case stipulates fewer conditions and therefore yields a more general
factual scenario.  We cap the number of representative edge cases to $25$ per provision,
dropping the edge cases with the largest number of input values first.

Finally, each representative is passed to Claude Opus~4.6 
together with the provision text, the two formalizations of
the witnessing pair, the fixed variables and their required truth
values, the remaining interface variables, and the root causes.  The
model returns a short description of a factual scenario.

%% file: sections/results.tex
\section{Results}
\label{sec:results}

\subsection{Matching Consistency}
\label{sec:results-matching-consistency}

\begin{figure}[t]
  \centering
  \includegraphics[width=\columnwidth]{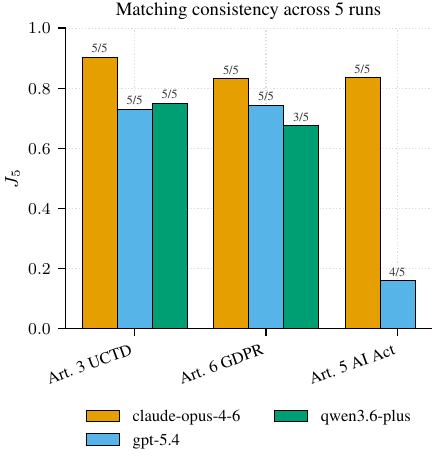}
  \caption{$J_5$ per (model, provision) across five independent
    matching runs.  $k/5$ above each bar denotes number of runs that returned
    a schema-valid matching. Provisions are ordered by estimated complexity
    from low to high.}
  \label{fig:consistency}
\end{figure}

We evaluate matching consistency by running each of three candidate
models five times on each of three provisions and measuring the
$N$-set Jaccard of the resulting matchings.
Each run is summarized by its \emph{co-membership pair set}
$\mathcal{P}_i$ --- the set of all atom pairs $\{(T, v),\,(T', v')\}$
placed in the same EC by run $i$, where each atom $(T, v)$ identifies
a node $v$ of formalization $T$.  The $N$-set Jaccard
\begin{equation}
  J_N = \frac{\displaystyle\Bigl|\bigcap_{i=1}^{N} \mathcal{P}_i\Bigr|}
             {\displaystyle\Bigl|\bigcup_{i=1}^{N} \mathcal{P}_i\Bigr|},
  \label{eq:n-set-jaccard}
\end{equation}
is the fraction of co-grouping decisions on which all $N$ runs agree.

Only Claude Opus~4.6 delivers consistent matchings on all three
provisions (Figure~\ref{fig:consistency}). GPT-5.4 produces
schema-valid output in all but one run but disagrees with itself on
Art.~5 AI~Act.  Qwen~3.6-plus already struggles on Art.~6 GDPR
($3/5$ valid runs) and fails entirely on Art.~5 AI~Act, the most
complex provision tested. Note that for a group with $k < 5$ valid runs, 
$J_N$ is computed over the valid runs only, so a
group is scored on a smaller intersection. 
The scores of the weaker matchers are optimistic in this respect.
We therefore adopt Claude Opus~4.6 for the matching step in all subsequent experiments.

\subsection{Edge Case Analysis}
\label{sec:results-edge-case-analysis}

\begin{figure*}[t]
  \centering
  \includegraphics[width=\textwidth]{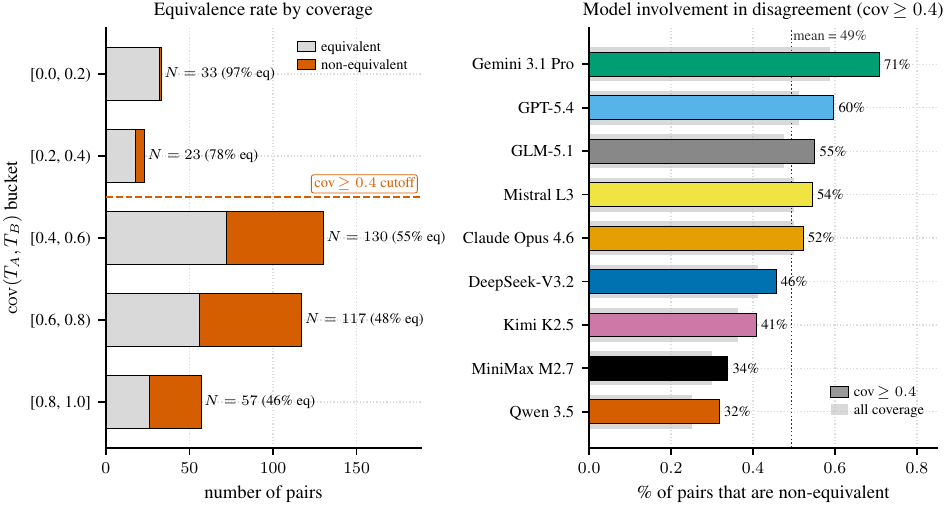}
  \caption{\emph{Left:} Share of equivalent and non-equivalent formalization pairs per 
    coverage bucket. Below the $\mathrm{cov} \geq 0.4$
    cutoff (dashed), pairs are dominated by trivial equivalences whose
    interface barely crosses either formalization.
    \emph{Right:} fraction of pairs involving each model that are
    non-equivalent for $\mathrm{cov} \geq 0.4$.
    Faded background bars repeat the same statistic without the
    coverage filter. Dotted line: overall mean across all pairs with $\mathrm{cov} \geq 0.4$.}
  \label{fig:model_edginess}
\end{figure*}

We generate one formalization per model and legal provision, producing $90$ formalizations and $360$ pairs of formalizations. We then calculate the interface for every pair. Note that pairs are not independent data points. Each formalization enters eight of these pairs, and all nine formalizations of a provision are matched in a single step.  The numbers below should therefore be interpreted as complete counts over the comparisons in our data, not estimates from a random sample.

Figure~\ref{fig:model_edginess} (left) groups all pairs by their coverage, i.e.\ the share of the formalizations 
matched to each other and therefore available for edge-case analysis.
Below a coverage of $0.2$, $97\%$ of pairs are equivalent; below
$0.4$, still $89\%$.  This is the expected behavior: when both formalizations
share almost no structure, there is little Boolean substance left to disagree over, 
so equivalence becomes the default rather than evidence of conceptual alignment.
Above $\mathrm{cov} \geq 0.4$ the equivalence rate flattens near
$50\%$ and stops moving meaningfully with further coverage.  We therefore
set a coverage threshold of $0.4$ for the further analysis, leaving us with $N = 304$ pairs.

Figure~\ref{fig:model_edginess} (right) shows the fraction of its pairs that are
non-equivalent per model. The difference between models is large: Gemini~3.1~Pro disagrees
with its peers on $71\%$ of its pairs, Qwen~3.5 on only $32\%$.  
However, the rate of disagreement spans
at least $50$ percentage points across the ten provisions, with
several swinging from $0\%$ to $100\%$ (not shown).

\begin{figure}[!t]
  \centering
  \includegraphics[width=\columnwidth]{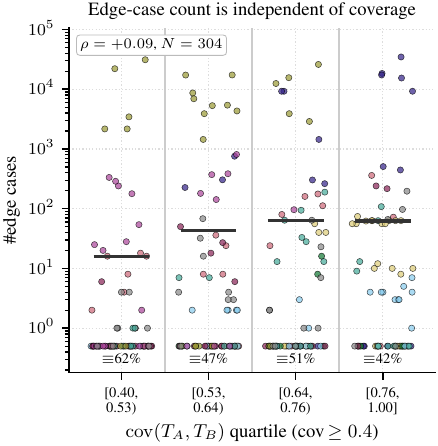}
  \caption{Edge-case count per pair across coverage quartiles.
    Dots are pairs, colored by provision. Dark ticks mark the median
    over non-equivalent pairs in each quartile.  The $\equiv\%$
    annotation reports the fraction of logically equivalent pairs per
    quartile. Coverage and edge-case count are almost unrelated
    across these pairs ($\rho = +0.09$).}
  \label{fig:pis_by_coverage}
\end{figure}

Figure~\ref{fig:pis_by_coverage} bins all pairs
into coverage quartiles and plots the edge-case count per pair.  The
spread is enormous --- three or more orders of magnitude in every bin
--- and the rank correlation between coverage and edge-case count is
negligible ($\rho = +0.09$).
The fraction of
equivalent pairs decreases with increasing coverage, but only slightly. Even at the highest-coverage end, $42\%$ of pairs are logically equivalent.
 Coverage thus characterizes \emph{when} a behavioral comparison is
informative, but not \emph{how much} the two formalizations actually
differ in their outcome. Even when models appear to agree on legal concepts, they can still diverge on many cases. Edge case analysis therefore adds a dimension of comparison that cannot be deduced from the matching information directly.

\subsection{Verbalization}
\label{sec:results-verbalization}

\begin{table*}[!t]
  \centering
  \footnotesize
  \setlength{\tabcolsep}{6pt}
  \renewcommand{\arraystretch}{1.25}
  \begin{tabular}{@{} p{0.10\textwidth} p{0.58\textwidth} p{0.24\textwidth} @{}}
    \toprule
    \textbf{Provision} & \textbf{Scenario \& question (LLM verbatim)} & \textbf{Model partition} \\
    \midrule
    Art.~5 AI~Act
    &
    SecureBio~GmbH places on the EU market a biometric categorisation system that individually
    categorises natural persons based on their facial geometry to infer their race. The system is
    marketed specifically for use by national police forces to categorise biometric data in the
    area of law enforcement. The system is not used for labelling or filtering of lawfully
    acquired biometric datasets. SecureBio also markets the system as intended for medical
    reasons, specifically to assist forensic pathologists in identifying genetic predispositions
    linked to racial phenotypes for post-mortem medical examinations.
    \smallskip\par
    \textit{Is SecureBio's biometric categorisation system a prohibited AI practice under
    Article~5?}
    &
    \textbf{Yes, prohibited (4):}
    \par
    Claude Opus 4.6, Kimi K2.5, MiniMax M2.7, Mistral L3
    \smallskip\par
    \textbf{No, not prohibited (4):}
    \par
    Gemini 3.1 Pro, GPT-5.4, GLM-5.1, Qwen 3.5
    \\
    \midrule
    Art.~3 UCTD
    &
    Maria joined FitLife Gym and signed a membership contract containing a clause that limits
    the gym's liability for personal injuries to \texteuro 500, regardless of fault. The
    clause was composed during the sign-up meeting by the gym manager using improvised
    language rather than being taken from a template or pre-existing document. Maria had no
    opportunity to influence or modify the substance of this clause. The gym manager cannot
    provide evidence that the liability cap was individually negotiated with Maria. The
    clause shifts substantially all injury risk onto Maria while relieving the gym of
    obligations that would otherwise arise under general contract law.
    \smallskip\par
    \textit{Should this contractual term be regarded as unfair under Article~3?}
    &
    \textbf{Yes, unfair (2):}
    \par
    Claude Opus 4.6, GLM-5.1
    \smallskip\par
    \textbf{No, not unfair (1):}
    \par
    Kimi K2.5
    \\
    \bottomrule
  \end{tabular}
  \caption{Two exemplary verbalized edge cases. The third
    column lists, for each side of the disagreement, the models
    whose formalization yields that conclusion on this fact pattern;
    models not listed produce a formalization that does not exhibit
    the relevant fork at the root cause and therefore do not
    contribute to either camp.}
  \label{tab:verbalizations}
\end{table*}

We apply the selection pipeline described in Section~\ref{methods-verbalization} and obtain $148$ representative edge cases across the ten provisions. Table~\ref{tab:verbalizations} shows two exemplary cases. The text of the provisions and visualizations of the formalizations can be found in Appendix~\ref{sec:appendix-case-studies}.

The first case demonstrates how an edge case can surface a genuine
interpretive tension in the underlying statute. Eight frontier LLMs
partition evenly on a highly sensitive provision concerning biometric
racial categorization. Inspecting the formalizations, the disagreement
traces to the structure of the carve-out clause: four models read
\textit{``categorizing of biometric data in the area of law enforcement''}
as a free-standing exception that exempts any law-enforcement use,
while the other four require the data to additionally have been
lawfully acquired. The carve-out's syntax is in fact contested
in the legal commentary: the European Commission's Guidelines
treat law enforcement as one illustrative application of a unified
exception \cite{eu_commission_2025_guidelines}, whereas civil-society
analyses argue that the statutory wording supports a narrower
reading \cite{algorithmwatch_2025_statement}. We take no position on
the correct interpretation; what matters for our purposes is that
the pipeline localized the disagreement on a fact pattern where
a legal expert can immediately recognize the underlying issue,
rather than leaving it buried inside structurally similar trees.

The second case illustrates a disagreement that
traces to an encoding error in one of the
formalizations. Kimi~K2.5 treats two conditions as both
necessary for a term to count as not individually negotiated
--- the term must be pre-drafted \emph{and} the supplier
must fail to discharge the burden of proof --- whereas
Claude Opus and GLM-5.1 allow either condition to suffice.
The latter reading is correct: Article~3(2)'s burden-of-proof
rule is a procedural safeguard, not an additional substantive
requirement, so Kimi's conjunctive encoding over-constrains
the definition.

However, this case also shows a limit of edge-case analysis. The pipeline only surfaces what the formalizations
\emph{disagree} on, while errors they share stay invisible. Here
both GLM-5.1 and Kimi encode the pre-drafting test of Article~3(2) 
as the \emph{definition} of ``not individually negotiated'',
missing that Article~3(2) only establishes a default case
and does not exclude other routes to the same
conclusion \cite[\S\,1.2.2]{ec2019uctd}.

%% file: sections/conclusion.tex
\section{Discussion}
\label{sec:discussion}
Conceptual similarity does not reduce disagreement
between formalizations, and higher coverage does not reduce it
either. Equivalence
checking is therefore a separate dimension of analysis that
matching alone cannot capture.  This finding is consistent with \citet{graus2026legal}, who also reports cases of high structural similarity, but low outcome similarity.

Second, verbalization of edge cases exposes qualitatively distinct types of disagreement in a way a legal expert can adjudicate or at least localize at a glance. However, edge-case analysis and verbalization do not yield the full picture. Not all structural differences lead to a final difference in outcome; they might even cancel each other out. Further, a formalization that yields the legally correct result for a certain case is not necessarily structurally correct, as the Art.~3 UCTD case illustrates.

The method has many potential applications beyond analysis of LLM-generated formalizations. It may be applied, e.g.\ to compare human-authored encodings of the same statute, or one statute before and after a legislative amendment to discover unintended consequences.

During development of the method, we initially attempted to compute the interface jointly across all $N$ formalizations, but quickly learned that this erases the very disagreements we want to study. Switching to pairwise SAT analysis, combined with the subsequent signature-based grouping, lets us make statements about $N$ formalizations while preserving each pair's full internal structure.

In ongoing work we seek to integrate this method in an iterative LLM-driven formalization loop, using the verbalizations as intermediate feedback steps to gradually improve the quality of LLM formalizations in a given context.

\section{Conclusion}
\label{sec:conclusion}

We presented a method for systematically comparing formalizations of law by the decisions they encode. Given multiple formalizations of the same provision, we match them into an equivalence-class graph, derive a shared interface for each pair, and use SAT to enumerate
the edge cases on which the two formalizations disagree. Selected edge cases are then verbalized into concrete factual scenarios. 
Applied to nine LLM-generated formalizations of ten EU provisions, we find that structural alignment does not predict behavioral divergence and that the verbalized cases expose qualitatively distinct types of disagreement, including divergences that mirror genuine controversies in the legal commentary.